\documentclass[conference]{IEEEtran}
\IEEEoverridecommandlockouts
\usepackage{cite}
\usepackage{amsmath,amssymb,amsfonts}
\usepackage{graphicx}
\usepackage{textcomp}
\usepackage{xcolor}
\usepackage{float}
\usepackage{algorithm}
\usepackage{algpseudocode}
\usepackage{url}
\usepackage{multirow}
\usepackage{hyperref}
\usepackage{graphicx}
\def\BibTeX{{\rm B\kern-.05em{\sc i\kern-.025em b}\kern-.08em
    T\kern-.1667em\lower.7ex\hbox{E}\kern-.125emX}}
\begin{document}

\title{Cross-Modal Computational Model of Brain-Heart Interactions via HRV and EEG Features\\
{\footnotesize \textsuperscript{*}}
\thanks{}
}

\author{\IEEEauthorblockN{1\textsuperscript{st} Malavika Pradeep}
\IEEEauthorblockA{
\textit{Digital University Kerala}\\
Thiruvananthapuram, India \\
malavikapradeep2001@gmail.com \\
ORCID:0009-0005-9215-3542}
\and
\IEEEauthorblockN{2\textsuperscript{nd} Akshay Sasi}
\IEEEauthorblockA{
\textit{Digital University Kerala}\\
Thiruvananthapuram, India \\
akshaysasi12.knr@gmail.com \\
ORCID: 0009-0009-3708-554X}
\and
\IEEEauthorblockN{3\textsuperscript{rd} Nusaibah Farrukh}
\IEEEauthorblockA{
\textit{Digital University Kerala}\\
Thiruvananthapuram, India \\
nusaibah.farrukh@gmail.com \\
ORCID: 0009-0004-1237-2693}
\and
\IEEEauthorblockN{4\textsuperscript{th} Rahul Venugopal}
\IEEEauthorblockA{ 
\textit{Centre for Consciousness Studies, NIMHANS}\\
Bangalore, India \\
rhlvenugopal@gmail.com \\
ORCID: 0000-0001-5348-8845}
\and
\IEEEauthorblockN{5\textsuperscript{th} Elizabeth Sherly}
\IEEEauthorblockA{
\textit{Digital University Kerala}\\
Thiruvananthapuram, India \\
sherly@duk.ac.in \\
ORCID: 0000-0001-6508-950X}
}

\maketitle

\begin{abstract}
The electroencephalogram (EEG) has been the gold standard for quantifying mental workload; however, due to its complexity and non-portability, it can be constraining. ECG signals, which are feasible on wearable equipment pieces such as headbands, present a promising method for cognitive state monitoring. This research explores whether electrocardiogram (ECG) signals are able to indicate mental workload consistently and act as surrogates for EEG-based cognitive indicators.

This study investigates whether ECG-derived features can serve as surrogate indicators of cognitive load, a concept traditionally quantified using EEG. Using a publicly available multimodal dataset (OpenNeuro) of EEG and ECG recorded during working-memory and listening tasks, features of HRV and Catch22 descriptors are extracted from ECG, and spectral band-power with Catch22 features from EEG. A cross-modal regression framework based on XGBoost was trained to map ECG-derived HRV representations to EEG-derived cognitive features. In order to address data sparsity and model brain-heart interactions, we integrated the PSV-SDG to produce EEG-conditioned synthetic HRV time series.
This addresses the challenge of inferring cognitive load solely from ECG-derived features using a combination of multimodal learning, signal processing, and synthetic data generation.

In this research, a cross-modal regression pipeline is constructed that projects ECG features into an EEG representative cognitive space; integration of a physiologically inspired synthetic HRV generator to augment cognitive-state modelling; and quantitative evidence that autonomic signals encode partial markers of mental workload. This brings a preliminary step toward bridging autonomic and central nervous system signals, highlighting both the potential and current limitations of ECG-based cognitive state inference.
These outcomes form a basis for light, interpretable machine learning models that are implemented through wearable biosensors in non-lab environments. Synthetic HRV inclusion enhances robustness, particularly in sparse data situations.
Overall, this work is an initiation for building low-cost, explainable, and real-time cognitive monitoring systems for mental health, education, and human-computer interaction, with a focus on ageing and clinical populations.
\end{abstract}


\section{Introduction}
Understanding the dynamics of brain and heart interaction during cognitive activities is now a developing area in both neuroscience and physiological computing. Early studies on the autonomic nervous system showed how complex networks of sympathetic and parasympathetic control may influence the heart through activity from the brain. As technologies improved, techniques such as electroencephalography and electrocardiography enabled the capture and analysis of these signals with remarkable acuity; each provides a very different window into the inner workings of human physiology. Central nervous activity is investigated in this study area to determine how it modulates heart rate variability.

EEG has traditionally been considered the gold standard for monitoring mental workload, memory, and attention because of its direct insight into brain dynamics. Meanwhile, HRV calculated from ECG became a reliable index of autonomic function associated with stress, fatigue, and emotional regulation. Yet, despite decades of research, EEG and ECG have mostly been studied in isolation. Only recently has the scientific community begun to explore their bidirectional relationship and how this interplay might reveal deeper insights into cognition and mental health.

Despite the powerful diagnostic capabilities of EEG, its practical limitations-which include high setup complexity, susceptibility to noise, and poor portability-make everyday environmental applications complex. As cognitive-monitoring applications increasingly extend toward mental health, workplace assessment, education, and human-computer interaction, there is increasing demand for lightweight, low-cost solutions that are deployable via consumer-grade wearable devices. ECG satisfies these practical constraints but has conventionally been regarded as inadequate for capturing higher-order cognitive states, as it reflects mainly autonomic modulation rather than direct neural activity.

This has set up a critical disconnect in that existing methods of monitoring the cognitive state depend mostly on EEG, whereas signals compatible with real-world use in scalable systems should be captured non-invasively using wearables. A question of outstanding scientific and engineering interest is whether ECG-derived features can serve as reliable surrogates for the EEG-based representations of cognition. Specifically, it remains unknown to what degree the autonomic responses reflected in the HRV encode information that correlates with the central neural processes underlying working memory and attention.

In this work, we explore a computational framework that integrates interpretable ECG feature engineering with multimodal learning and synthetic HRV generation to estimate EEG-like cognitive representations from ECG data alone. The goal is to determine if autonomic signal features can be mapped into a space that meaningfully approximates EEG-derived cognitive markers and thus allows for light-weight classification of cognitive state. The key innovations introduced are as follows:

 1.	ECG and EEG Feature Engineering: The informational characteristics of the ECG signals were extracted using HRV measures in the time domain, and Catch22 time-series features. Similarly, EEG features were derived, representing spectral band power and time-series behaviour to provide a robust cognitive state representation.
\newline 2. Cross-modal learning: The key novelty here is the application of a regression model using XGBoost to project HRV features from ECG onto cognitive features from EEG. This cross-modal learning bridges representations across the autonomic and central nervous systems to classify mental states with ECG.
\newline 3.	Synthetic Data Augmentation through PSV-SDG: PSV-SDG is employed for the generation of realistic HRV signals, the modulating factors being given by the brain activities. This, in fact, increases model strength, especially when real data is noisy or limited.

The work, through intensive experimentation in publicly released multimodal databases, shows that the features extracted from ECG, when processed using the suggested pipeline, can predict the representations of EEG with comparable accuracy and efficiently identify cognitive states such as passive listening and memory load \\cite{b7}. It is also highlighted that the employment of synthesized HRV data increases robustness and generalizability, supporting the cross-modal strategy\\cite{b6}. These results are important as they dispel the long-held notion that EEG is necessary to detect cognitive state. By establishing ECG as a valid proxy for certain cognitive indicators, the present study forms a basis for scalable, affordable, and real-world solutions for mental state monitoring compatible with wearables already in mass usage.

\section{Related Work}
Recent studies highlight EEG and HRV as promising non-invasive biomarkers for Mild Cognitive Impairment (MCI), where EEG provides insights into neural oscillations and HRV reflects autonomic regulation, though each alone achieves moderate accuracy. Multimodal approaches, such as combining EEG and HRV features during CERAD tasks with hybrid machine learning models, have significantly improved detection performance, reaching accuracies up to 97\%, surpassing unimodal and traditional methods\cite{b6}. Candia-Rivera and Chavez (2025)\cite{b1} introduced a framework for dyadic cardiac rhythmicity analysis, revealing bilateral maternal–fetal cardiac couplings using time-varying Poincaré-based measures. Their work highlighted that fetal heart activity may play an active signaling role in co-regulatory mechanisms, with stronger coupling observed during labor, underscoring the autonomic system’s complex bidirectional communication.

A Brain Sciences study (2022) demonstrated that HRV correlates with EEG activity during working memory tasks in patients with depressive and anxiety disorders.\cite{b5}

Methodological advances have also been proposed. For instance, the MethodsX paper \cite{b3} detailed computational pipelines for the joint analysis of EEG and ECG signals in memory-related tasks. These works collectively demonstrate that HRV not only reflects autonomic regulation but also serves as a surrogate marker of neural processing, particularly when combined with EEG-based features. 

Works related to the computational model "Functional assessment of bidirectional cortical and peripheral neural control on heartbeat dynamics"\cite{b4} talk about a novel computational framework SV-SDG, that functionally assesses bidirectional brain-heart interplay, combining cortical activity of EEG and peripheral neural dynamics derived from heart rate variability. 

Another work, "Analysis of EEG and ECG time series in response to olfactory and Cognitive tasks"\cite{b11} investigates the correlation between heart and brain activity when stimulated by olfactory stimuli and cognitive activity using EEG and ECG signals finding significant correlation between the signals under various conditions, including rest and aromatherapy.

\section{Methodology}
This study applies the publicly posted OpenNeuro Dataset ds003838 entitled "EEG, cardiovascular, and pupillometry responses during memory load tasks". The dataset is available on: \url{https://openneuro.org/datasets/ds003838}.
The data were selected due to their multimodal nature, as they consist of synchronized recordings of EEG, ECG, photoplethysmography (PPG), and pupillometry signals, recorded from 86 healthy adult subjects. Recordings feature a clearly defined cognitive task paradigm with different levels of memory load (digit span tasks with 5, 9, or 13 digits) and passive listening conditions. This is a perfect basis for analyzing brain-heart interactions and the applicability of ECG-based cognitive state estimation.
Data Usage Statement:
The dataset is governed by the OpenNeuro Terms of Use, which allows free reuse under CC0 Public Domain Dedication. Ethical clearance was managed by the authors of the original dataset.

\subsection{Working Model}

The end-to-end pipeline of EEG and ECG signal processing is described in Fig. 1. The process starts with raw data collected from Openneuro, followed by individual pre-processing of EEG and ECG modalities. The signals are transformed into a structured five-dimensional array and stored in pickle format. Feature extraction is performed to extract catch-22 features of EEG and ECG and HRV features of ECG. Synthetic HRV features are generated using the Poincaré Sympathetic-Vagal Synthetic Data Generation (PSV-SDG) model to enrich the dataset. Machine learning classifiers, such as XGBoost and Random Forest, and convolutional neural networks, are trained and tested for binary and multiclass classification tasks.
\begin{figure}[H]
\includegraphics[width=0.5\textwidth]{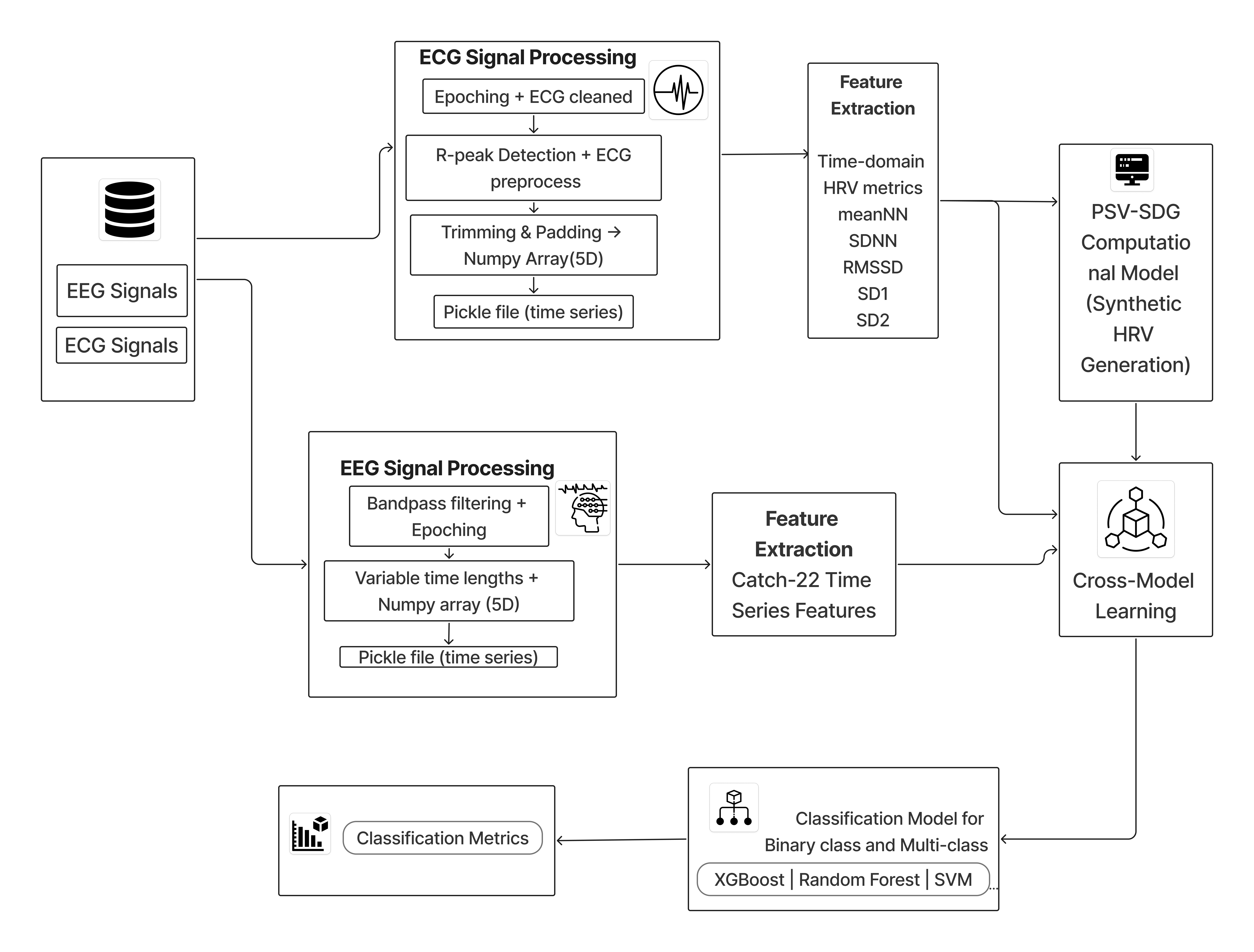}  
  \caption{Workflow of the proposed cross-modal framework that combines ECG and EEG signals: it includes preprocessing, extraction of HRV and Catch22 features, generation of synthetic HRV by the PSV-SDG model, and cross-model learning using interpretable and low-computation classifiers such as Random Forest, XGBoost, and SVM.}
  \label{fig:task}
\end{figure}

\subsection{Data Acquisition}
Throughout the experiment, every subject performed a digit-span memory task while physiological signals such as 64-channel EEG, ECG, PPG, and pupillometry were recorded in Brain Imaging Data Structure (BIDS) format (Fig.2). \cite{b7}
\subsubsection{Experimental Paradigm}
As in the Fig. 2., all trials consisted of successive phases: (i) a 0.5-second warning stimulus, (ii) a 1-second cue label expressing "Just Listen" or "Memorize," (iii) a 3-second fixation baseline, (iv) auditory presentation of sequences of digits (five, nine, or thirteen digits), (v) a recall period for the memory conditions (7 to 15 seconds depending upon the load), and (vi) a 5-second trial-to-trial interval. Event markers embedded within physiological recordings synchronized the task phases with neural and cardiovascular responses.
The dataset has the potential to examine the memory-related brain-heart dynamics. It enables the analysis of electroencephalographic systems (ERPs, oscillatory activity) and cardiovascular systems (ECG, PPG, and HRV), as well as their interactions and dynamics under varying levels of cognitive load.
\begin{figure}[H]
\includegraphics[width=0.4\textwidth]{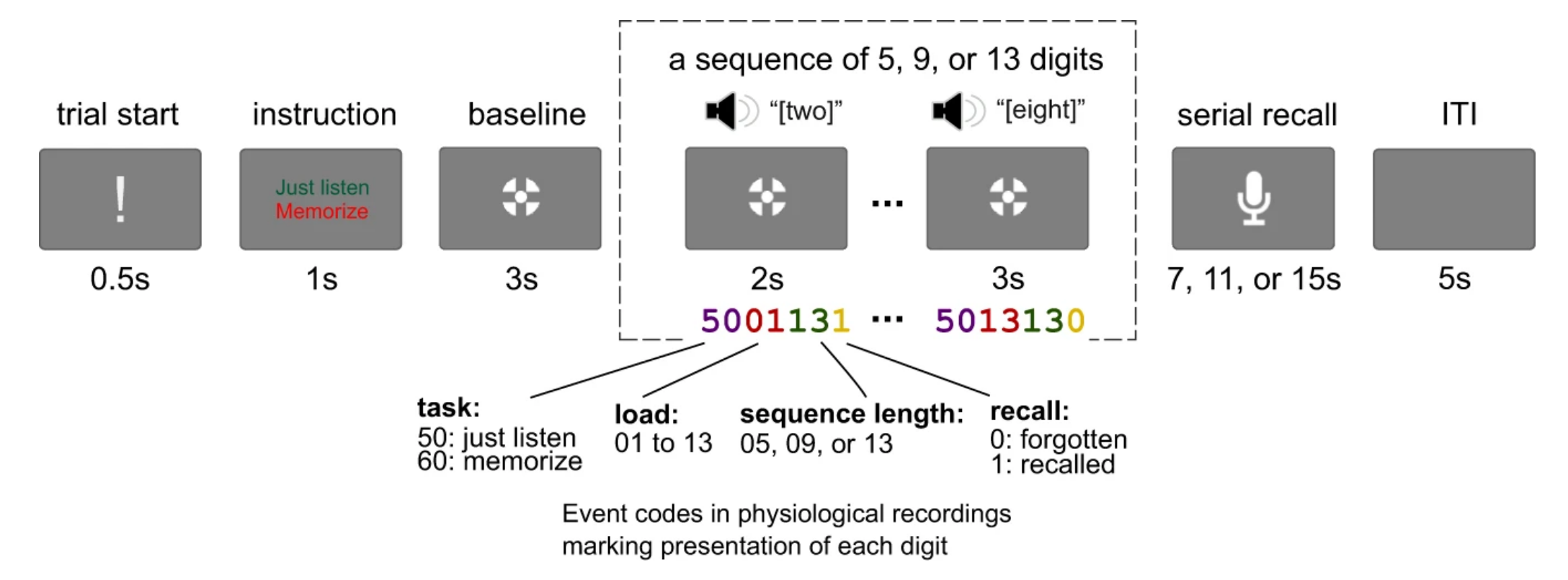}  
  \caption{Experimental design of the working memory task: trial structure with instruction, baseline, presentation of digit sequence (5, 9, or 13 digits), serial recall, and inter-trial interval is shown.}
  \label{fig:task}
\end{figure}

\subsection{Signal Pre-processing}
\subsubsection{ECG Preprocessing}
Using MNE-Python, extracted ECG signals from .set files as part of the pre-processing step. Event markers indicated the time intervals for epoching from -3s to the time requirements for the tasks. R-peak identification and noise elimination, as well as the extraction of the ECG signal, were accomplished with NeuroKit2. Each epoch, which was elongated to a maximum trial duration, was normalized and arranged in a 4D format in terms of the subject and then consolidated into a 5D array structure (subjects, conditions, subconditions, trials, time). The preprocessed 5D arrays were stored as pickle files.
\subsubsection{EEG Preprocessing}
The channels that were kept during EEG preprocessing were aligned to the 10–20 system. Signals underwent bandpass filtering (1–40 Hz) and were re-referenced to the average reference. Epoching was done relative to the event markers ( -3 s pre-stimulus to the end of the task). Baseline correction was performed on pre-stimulus intervals, along with rejection of high-amplitude artifacts. Surviving epochs length normalized and organized into 5D structure ([subjects, conditions, subconditions, trials, channels, time]).

\subsection{Feature Extraction}
Catch22 was used to extract ECG time-series features from each trial. The 22 interpretable measures included autocorrelation, stationarity, entropy, and nonlinearity. Storage of extracted features included trial-specific condition and subcondition labels.
EEG time-series(memory specific trials) were standardized and processed by catch22. Organized dataframes were stored for each subject, channel, and trial feature set.
HRV metrics were calculated from R-peak detection RR intervals. MeanNN, SDNN, RMSSD, and Poincaré descriptors (SD1, SD2) were time-domain features. All feature sets were indexed by subject, condition, and trial.

\subsection{PSV-SDG Framework: Computational Modelling} 
The Poincaré Sympathetic-Vagal Synthetic Data Generation (PSV-SDG) model is used to simulate brain–heart interactions by integrating EEG activity with autonomic nervous system (ANS) dynamics\cite{b3}. Short-term and long-term variability indices were computed from inter-beat intervals (IBI) using:
\begin{align}
SD_{01} &= \frac{1}{2} \text{std}(IBI')^2 \\
SD_{02} &= 2\text{std}(IBI)^2 - \frac{1}{2} \text{std}(IBI')^2 \\
SD_1(t) &= \frac{1}{2} \text{std}(IBI'_{\Omega_t})^2 \\
SD_2(t) &= 2\text{std}(IBI_{\Omega_t})^2 - \frac{1}{2} \text{std}(IBI'_{\Omega_t})^2
\end{align}

Synthetic RR intervals were generated using the Integral Pulse Frequency Modulation (IPFM) model:

\begin{equation}
x(t) = \sum_{k=1}^{N} \delta(t - t_k)
\end{equation}

\begin{equation}
1 = \int_{t_k}^{t_{k+1}} [\mu_{HR} + m(t)] dt
\end{equation}
with modulation signal:
\begin{equation}
m(t) = C_S \sin(\omega_S t) + C_V \sin(\omega_V t),
\end{equation}
where $\omega_S=2\pi \cdot 0.1$ Hz and $\omega_V=2\pi \cdot 0.25$ Hz represent sympathetic and vagal oscillations.

\subsubsection{Synthetic HRV Generation:}
ECG trials were cleaned, R-peaks detected, and RR intervals extracted. Using a 30-beat sliding window, dynamic $SD_{1}$ and $SD_{2}$ values were computed and used in the PSV-SDG model to synthesize HRV signals. Approximately 1500 synthetic trials were generated across subjects and conditions. 
The synthetic HRV signals created with the PSV-SDG model were added to the computational framework, which augments the data with generated HRV signals for robustness to support real ECG-derived features.
This synthetic method tackles issues in real-world HRV acquisition, such as noise, variability, and insufficient data. It also allows for controlled simulation of autonomic dynamics. In addition to improving data quantity, it aids in testing hypotheses and validating algorithms, leading to stronger and more general models for analyzing brain-heart interactions.

\subsection{Evaluation and Setup}
The evaluation was performed to test unimodal and cross-modal classification strategies. For model training, hyperparameters were optimized: 

\setlength{\tabcolsep}{10pt}
\begin{table}[ht]
\centering
\label{tab:hyperparams}
\begin{tabular}{|l|p{5.5cm}|}
\hline
\textbf{Model} & \textbf{Hyperparameters} \\
\hline
Random Forest & 
n\_estimators $= 500$, 
max\_depth $= 10$, 
max\_features = $\sqrt{\phantom{x}}$, 
min\_samples\_leaf $= 2$, 
min\_samples\_split $= 10$, 
bootstrap = False, 
class\_weight = balanced, 
random\_state $= 42$, 
n\_jobs = -1 \\
\hline
SVM & 
kernel = RBF, 
class\_weight = balanced, 
random\_state $= 42$ \\
\hline
XGBoost & 
n\_estimators=500, 
max\_depth= 6, 
learning\_rate=0.1, 
objective= multi:softmax, 
num\_class=3, 
random\_state $= 42$ \\
\hline
\end{tabular}
\vspace{0.5em}
\caption{Hyperparameter settings for classification models}
\end{table}
Unimodal Classification: Random Forest (RF), Support Vector Machine (SVM), and XGBoost (XGB) were trained separately on HRV features, Catch22 ECG features, synthetic HRV features, combined HRV + synthetic HRV, and EEG features. Both multiclass (Five, Nine, Thirteen memory loads) and binary (JustListen vs. Memory) tasks were considered.

Cross-Modal Classification: A cross-model regression was trained (XGBoost) to evaluate ECG as a substitute of EEG, which maps HRV features (meanNN, SDNN, RMSSD, SD1, SD2) onto EEG catch22 features. The regressed features were classified using EEG-trained models (Random Forest, XGBoost, CNN), which predict the cognitive load. Class imbalance is addressed using SMOTE, and performance is evaluated using the F1 score, accuracy, and confusion matrix.

Evaluation Metrics: Accuracy, precision, recall, and F1-score were used. For imbalanced classes, SMOTE oversampling was applied.
\subsection{Results}
In unimodal classification, the performance of HRV-based models varied substantially. Using time-domain HRV features, multiclass prediction of Five, Nine, and thirteen- memory loads achieved moderate accuracy, with Random Forest (0.81) followed by XGBoost(0.79) and SVM(0.69). In contrast, the binary classification Just Listen v/s Memory task yielded weak performance, with XGBoost having the highest accuracy - 0.56. Catch22 ECG features produced a different pattern; with XGBoost reaching 0.99 accuracy, Random Forest (0.98), and SVM trailing at 0.89. Their utility did not extend to binary classification, where classifiers dropped to less than 0.59.

Synthetic HRV features generated by the PSV-SDG model demonstrated strong discriminative ability. In multiclass, XGBoost attained 0.97 accuracy with F1 scores of 0.94, while Random Forest(RF) attained 0.90; SVM performed poorly (0.35). Binary classification remained difficult with RF and SVM accuracies near 0.54 and no stable result for XGBoost. When HRV and synthetic HRV features were combined, RF achieved the highest multiclass performance (0.93, F1=0.92-0.96), whereas SVM deteriorated (0.34). Binary prediction agin remained weak across all models (0.53-0.55). 
EEG-derived features resulted in almost perfect multiclass classification performance for all models (XGBoost 0.998, Random Forest 0.996, SVM 0.974), binary classification remained limited, with accuracy between 0.55 and 0.61 and F1 scores below 0.74.

A statistical comparison of classifiers using one-way ANOVA on 5-fold cross-validation accuracies confirmed significant performance differences in multiclass classification, F(2,12)= 378.55, p < 0.0001. Turkey post-hoc analysis revealed the Random Forest(M= 0.810) outperformed both XGBoost(M=0.790, p = 0.0106) and much better than SVM (M= 0.670, p < .001), reinforcing the relative advantage of ensemble methods for HRV-based cognitive-state prediction. For binary task, XGBoost (M=0.6662) followed by RF(M=0.604).

In cross-modal experiments, mapping HRV features into EEG-derived space through regression and subsequent classification resulted in an accuracy of 40\%, only modestly above the chance level of 33\%. Class-wise analysis showed that Thirteen-memory was the most identifiable class (recall 0.60, F1 0.49), whereas Nine-Memory was poorly represented. These limitations primarily reflect the loss of EEG-specific variance during regression and increased noise in the reconstructed features, rather than the shortcomings of the classification model.

\begin{table}[H]
\centering
\vspace{0.5em}
\resizebox{\linewidth}{!}{%
\begin{tabular}{|l|c|c|c|c|c|c|}
\hline
\textbf{Feature Set} & \multicolumn{3}{c|}{\textbf{Multiclass}} & \multicolumn{3}{c|}{\textbf{Binary}} \\
\hline
 & RF & SVM & XGB & RF & SVM & XGB \\
\hline
HRV Features       & 0.81  & 0.69  & 0.79  & 0.503 & 0.525 & \textbf{0.556} \\
Synthetic HRV      & 0.903 & 0.353 & \textbf{0.969} & 0.539 & \textbf{0.548} & 0.546 \\
Combined Synthetic & \textbf{0.933} & 0.342 & 0.342 & 0.532 & 0.548 & 0.548 \\
EEG Features       & 0.996 & 0.974 & \textbf{0.998} & 0.555 & \textbf{0.609} & 0.575 \\
\hline
\end{tabular}
}
\vspace{0.5em}
\caption{Test Accuracies of Different Feature Sets for Multiclass and Binary Classification}
\label{tab:classification_results}
\end{table}

\begin{table}[H]
\centering
\vspace{0.5em}
\resizebox{\linewidth}{!}{%
\begin{tabular}{|l|l|c|c|c|}
\hline
\textbf{Classifier} & \textbf{Class} & \textbf{Precision} & \textbf{Recall} & \textbf{F1-Score} \\
\hline
\multirow{3}{*}{XGBoost} 
  & Five-Memory     & 0.33 & 0.45 & 0.38 \\
  & Nine-Memory     & 0.38 & 0.08 & 0.14 \\
  & Thirteen-Memory & \textbf{0.41} & \textbf{0.60} & \textbf{0.49} \\
\hline
\multirow{3}{*}{Random Forest} 
  & Five-Memory     & 0.33 & 0.45 & 0.38 \\
  & Nine-Memory     & 0.38 & 0.08 & 0.14 \\
  & Thirteen-Memory & \textbf{0.41} & \textbf{0.60} & \textbf{0.49} \\
\hline
\multicolumn{2}{|l|}{\textbf{Overall Accuracy}} & \multicolumn{3}{c|}{0.40} \\
\hline
\multicolumn{2}{|l|}{\textbf{Macro Average}}    & 0.42 & 0.40 & 0.33 \\
\hline
\multicolumn{2}{|l|}{\textbf{Weighted Average}} & 0.42 & 0.40 & 0.33 \\
\hline
\end{tabular}
}
\vspace{0.5em}
\caption{Performance Metrics of XGBoost and Random Forest for Cross-Model HRV-to-EEG Classification}
\label{tab:cross_model_results}
\end{table}

\subsection{Summary}
In this research, state classification has been investigated with both cross-modal and unimodal approaches. HRV features (meanNN, SDNN, RMSSD, SD1, SD2) and Catch22 features were extracted from the ECG, while synthetic HRV signals were generated using the PSV-SDG model. Features of EEG retaining spectral and nonlinear dynamics were also extracted. Random Forest, Support Vector Machine, and XGBoost classifiers were applied across multiclass (Five, Nine, Thirteen memory loads) and binary (JustListen vs. Memory) tasks. Cross-modal mapping applied XGBoost regression projecting HRV onto EEG feature space, with classification being assessed using accuracy, F1-score, and other respective metrics.

\section{Discussion and Conclusion}
This work investigated cross-modal classification of cognitive states between EEG and ECG signals with emphasis on identifying various memory load states and passive listening states. Having extracted informative time-series features such as Catch-22 descriptors from EEG and heart rate variability measures from ECG, we trained different machine learning classifiers such as Random Forest, SVM, and XGBoost for
binary and multiclass discrimination. While multiclass memory-load prediction using ECG features showed moderate performance, the cross modal mapping from HRV to EEG achieved marginal accuracy . This limited accuracy clearly demonstrates that EEG and ECG capture largely orthogonal information about brain states. Because of this orthogonality, using either modality in isolation drives classifier performance closer to chance levels.
To further aid in modeling and generalization, we included the Poincar´e Sympathovagal Synthetic Data Generation (PSV-SDG) model, which produces physiologically realistic HRV signals by modulating synthetic RR intervals with inferred brain activity. The synthetic addition enhances the richness and variability of the ECG signal, further enabling stronger learning and evaluation. These findings suggest that while ECG alone is currently still short of EEG's discriminability, its ease of use with wearable sensors and noninvasive nature position it as an appealing signal source for scalable, multimodal brain monitoring systems.

\section{Limitation and Future Work}
The relatively low cross-modal accuracy is likely due to the physiological and temporal difference between EEG and HRV signals — EEG signals reflect fast cortical activity, while HRV captures slow autonomic patterns. Point-by-point regression of signals loses the important timing and spectral information, and sparse paired data diminishes generalizability even more. Since the objective is to create lightweight and interpretable models that can be used in real-time or wearable devices, future research could explore efficient options such as shallow neural networks, logistic regression, LightGBM, or ridge and LASSO regression, which entail minimal computational cost and maintain explainability through feature importance. Combining them with elementary alignment techniques like Canonical Correlation Analysis (CCA) or linear contrastive learning can achieve accuracy versus interpretability balancing. Besides, topic-wise normalization and synthetic-to-real domain adaptation can also enhance the reliability without making the system heavy computationally. 

With the advent of smart wearables, which include continuous heart rate, HRV, sleep, and metabolic biomarker monitoring, the proposed SV-SDG and EEG-ECG multimodal method can be designed to facilitate real-time cognitive load measurement outside of lab-controlled environments. While these rings lack EEG sensors, their high-fidelity activity and HRV data make them ideal platforms for the application of ECG-derived features and synthetic modelling (PSV-SDG) as antecedents to cognitive strain or fatigue.
Future work can be aimed at building lightweight implementations of the current model to run on wearable-friendly edge devices. Secondly, EEG data—traditionally involving laboratory-like settings—can be complemented by way of neuro-adaptive interfaces or supplemented with passive behavioural signals (e.g., typing speed, eye movement) to approximate cognitive state. This opens the door to real-time brain-heart interaction modelling with wearables alone, allowing for the tracking of mental workload in ecologically valid situations such as
workplaces, classrooms, or sports training.
The method can also contribute to personalized cognitive well-being tracking via accommodation to
individualized baselines and cycles, an ever more realistic aspiration with longitudinal wearables data. Built into
platforms like Ultrahuman, PSV-SDG might enable proactive cognitive wellness feedback, stress predictions, and
even intervention alerts (e.g., recommending breaks or mindfulness) — making this study an important next
step toward the next-generation of cognitive-aware wearable systems.

\section{Code Availability}
The code and experimental resources used in this study are publicly available at: https://github.com/Malavika-pradeep/Computational-model-for-Brain-heart-Interaction-Analysis

\section{Acknowlegement}
The authors would like to gratefully acknowledge the support of their advisors and faculty for their valuable guidance and the computational resources provided by the University's GPU server facility. The authors acknowledge the use of large language model tools for language polishing and improving the readability. No AI system contributed to the analysis, results or scientific claims.

\section*{}

\end{document}